\documentclass[runningheads]{llncs}
\usepackage{graphicx}
\usepackage{footmisc} % multi-footnotes
\usepackage[utf8]{inputenc}
\usepackage{amsmath}
\usepackage{hyperref,xcolor}
\begin{document}
\title{Rethinking Anticipation Tasks: Uncertainty-aware Anticipation of Sparse Surgical Instrument Usage for Context-aware Assistance\thanks{Funded by the German Research Foundation (DFG, Deutsche Forschungsgemeinschaft) as part of Germany’s Excellence Strategy – EXC 2050/1 – Project ID 390696704 – Cluster of Excellence “Centre for Tactile Internet with Human-in-the-Loop” (CeTI) of Technische Universität Dresden.}}
\titlerunning{Anticipation of Surgical Instruments}
\author{Dominik Rivoir\inst{1,3} \and
Sebastian Bodenstedt\inst{1,3} \and
Isabel Funke\inst{1,3} \and
Felix von Bechtolsheim\inst{2,3} \and
Marius Distler\inst{2,3} \and
Jürgen Weitz\inst{2,3} \and
Stefanie Speidel\inst{1,3}}
%
% index{Rivoir, Dominik}
% index{Bodenstedt, Sebastian}
% index{Funke, Isabel}
% index{von Bechtolsheim, Felix}
% index{Distler, Marius}
% index{Weitz, Jürgen}
% index{Speidel, Stefanie}
%
\authorrunning{D. Rivoir et al.}
\institute{Translational Surgical Oncology, National Center for Tumor Diseases (NCT),
Dresden, Germany\\ \email{dominik.rivoir@nct-dresden.de} \and
Department of Visceral, Thoracic and Vascular Surgery, Faculty of Medicine and
University Hospital Carl Gustav Carus, Technische Universität Dresden, Germany\and
Centre for Tactile Internet with Human-in-the-Loop (CeTI), TU Dresden, Germany}
\maketitle              % typeset the header of the contribution
\begin{abstract}
Intra-operative anticipation of instrument usage is a necessary component for context-aware assistance in surgery, e.g. for instrument preparation or semi-automation of robotic tasks. However, the sparsity of instrument occurrences in long videos poses a challenge. Current approaches are limited as they assume knowledge on the timing of future actions or require dense temporal segmentations during training and inference. We propose a novel learning task for anticipation of instrument usage in laparoscopic videos that overcomes these limitations. During training, only sparse instrument annotations are required and inference is done solely on image data. We train a probabilistic model to address the uncertainty associated with future events. Our approach outperforms several baselines and is competitive to a variant using richer annotations. We demonstrate the model's ability to quantify task-relevant uncertainties. To the best of our knowledge, we are the first to propose a method for anticipating instruments in surgery.

\end{abstract}
\section{Introduction}
\begin{figure}
\centering
%\frame{
\includegraphics[width=\textwidth ,trim=1cm .4cm 1cm 9cm, clip]{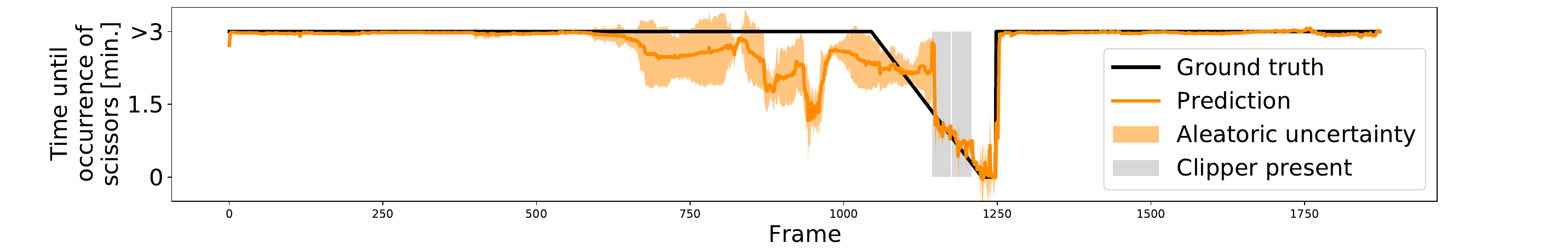}
%}
\caption{Predicted remaining time until occurrence of the scissors within a horizon of three minutes on an example surgery. In cholecystectomies, the clipper is an indicator for future use of the scissors. Error and uncertainty decrease when the clipper appears.} \label{fig:example}
\end{figure}
Anticipating the usage of surgical instruments before they appear is a highly useful task for various applications in computer-assisted surgery. It represents a large step towards understanding surgical workflow to provide context-aware assistance. For instance, it enables more efficient instrument preparation~\cite{maier2017surgical}. For semi-autonomous robot assistance, instrument anticipation can facilitate the identification of events that trigger the usage of certain tools and can eventually help a robotic system to decide when to intervene. Anticipating the use of certain instruments such as the irrigator can further enable early detection and anticipation of complications like bleedings.

The proposed applications require continuous anticipation estimates in long surgeries. Many instruments occur only rarely and briefly, i.e. only sparse annotations are available. Nevertheless, a useful anticipation framework should only react to these sparse occurrences and remain idle otherwise. Our approach addresses these requirements and is thus applicable to real-world scenarios. We train a neural network to regress the remaining time until occurrence of a specific instrument within a given future horizon. Our uncertainty-aware framework addresses the uncertainty associated with future events. This enables the identification of trigger events for instruments by measuring decreases in uncertainty associated with anticipation estimates (Fig. \ref{fig:example}).
Our task is inspired by predicting remaining surgery duration~\cite{twinanda2018rsdnet} but extends to arbitrary surgical events.

Various works have investigated short-horizon anticipation of human actions~\cite{damen2018scaling,gao2017red,jain2016recurrent,vondrick2016anticipating}. However, these methods are only designed to predict actions in the immediate future based on a single frame~\cite{damen2018scaling,vondrick2016anticipating} or short sequences of frames~\cite{gao2017red}. Most importantly, they are trained and evaluated on artificially constructed scenarios where a new action is assumed to occur within typically one second and only its correct category is unknown. In surgery however, the challenge rather lies in correctly timing predictions given the video stream of a whole surgical procedure. Our task definition considers long video sequences with sparse instrument usage and encourages models to only react to relevant cues. Jain et al.~\cite{jain2016recurrent} address this by adding the default activity \textit{'drive straight'} for anticipation of driver activities. For our task, we propose a similar category when no instrument is used within a defined horizon.

While methods for long-horizon anticipation exist, they require dense, rich action segmentations of the observed sequence during inference and mostly do not use visual features for anticipation~\cite{abu2019uncertainty,abu2018will,du2016recurrent,ke2019time,mehrasa2019variational}. This is not applicable to our task, since the information required for anticipating the usage of some instruments relies heavily on visual information. For instance, the usage of the irrigator is often triggered by bleedings and cannot be predicted solely from instrument signals. Some methods utilize visual features but nevertheless require dense action labels~\cite{mahmud2017joint,zhong2018time}. However, these labels are tedious to define and annotate and therefore not applicable to many real-world scenarios, especially surgical applications. In contrast, we propose to predict the remaining time until the occurrence of sparse events rather than dense action segmentations. During training, only sparse instrument annotations are required and inference is done solely on image data. This aids data annotation since instrument occurrence does not require complex definitions or expert knowledge~\cite{maier2014can}.

The uncertainty associated with future events has been addressed by some approaches~\cite{abu2019uncertainty,vondrick2016anticipating}. Similar to Farha et al.~\cite{abu2019uncertainty}, we do so by learning a probabilistic prediction model. Bayesian Deep Learning through Monte-Carlo Dropout provides a framework for estimating uncertainties in deep neural networks~\cite{gal2016dropout}. Kendall et al.~\cite{kendall2017uncertainties} identify the model and data as two relevant sources of uncertainty in machine learning. Several approaches have been proposed for estimating these quantities~\cite{gal2017deep,kwon2020uncertainty,shridhar2018uncertainty,wang2019aleatoric}. We evaluate our model's ability to quantify task-relevant uncertainties using these insights. The contributions of this work can be summarized as follows:
\begin{enumerate}
\item To the best of our knowledge, we are the first to propose a method for anticipating the use of surgical instruments for context-aware assistance.
\item We reformulate the general task of action anticipation to alleviate limitations of current approaches regarding their applicability to real-world problems.
\item Our model outperforms several baseline methods and gives competitive results by learning only from sparse annotations.
\item We evaluate our model's ability to quantify uncertainties relevant to the task and show that we can improve performance by filtering uncertain predictions.
\item We demonstrate the potential of our model to identify trigger events for surgical instruments through uncertainty quantification.
\end{enumerate}

\section{Methods}

\subsection{Anticipation Task} \label{sec:task}
\subsubsection{Regression:}Inspired by Twinanda et al.'s~\cite{twinanda2018rsdnet} remaining surgery duration task, we define surgical instrument anticipation as a regression task of predicting the remaining time until the occurrence of one of $K$ surgical instruments within a future horizon of $h$ minutes. Given frame $x$ from a set of recorded surgeries and an instrument $\tau$, the ground truth for the regression task is defined as
\begin{equation} \label{eq:definition}
r_h(x,\tau) = \min\{t_x(\tau),h\},
\end{equation}
where $t_x(\tau)$ is the true remaining time in minutes until the occurrence of $\tau$ with $t_{x'}(\tau)=0$ for frame $x'$ where $\tau$ is present. The target value is truncated at $h$ minutes since we assume that instruments cannot be anticipated accurately for arbitrarily long intervals. This design choice encourages the network only to react when the usage of an instrument in the foreseeable future is likely and predict a constant otherwise. Opposed to current definitions for anticipation tasks, we do not assume an imminent action or rely on dense action segmentations.

\subsubsection{Classification:}For regularization, we add a similar classification objective to predict one of three categories $c_h(x,\tau) \in \{anticipating,present,background\}$, which correspond to an instrument appearing within the next $h$ minutes, being present and a background category when neither is the case. In Section \ref{sec:uncert}, we discuss the benefits of this regularization task for uncertainty quantification.

\subsection{Bayesian Deep Learning}
Due to the inherent ambiguity of future events, anticipation tasks are challenging and benefit from estimating uncertainty scores alongside model predictions. Bayesian neural networks enable uncertainty quantification by estimating likelihoods for predictions rather than point estimates~\cite{kendall2017uncertainties}. Given data $X$ with labels $Y$, Bayesian neural networks place a prior distribution $p(\omega)$ over parameters $\omega$ which results in the posterior 
$p(\omega |Y,X) = p(Y|X,\omega)p(\omega)/\int p(Y|X,\omega)p(\omega)d\omega$. Since the integration over the parameter space makes learning $p(\omega|Y,X)$ intractable, variational inference~\cite{graves2011practical} is often used to approximate $p(\omega|Y,X)$ by minimizing the Kullback-Leibler divergence $KL(q_\theta(\omega)||p(\omega|Y,X))$ to a tractable distribution $q_\theta(\omega)$. Gal et al.~\cite{gal2016dropout} have shown that this is equivalent to training a  network with dropout if $q_\theta(\omega)$ is in the family of binomial dropout distributions.

During inference, we draw $T$=10 parameter samples $\omega_t \sim q_\theta(\omega)$ to approximate the predictive expectation $E_{p(y|x,Y,X)}(y^r)$ for regression variables $y^r$ and the predictive posterior $p(y^c|x,Y,X)$ for classification variables $y^c$ (Eq. \ref{eq:exp} \& \ref{eq:post}), where $f_{\omega_t}(x)$ and $p_{\omega_t}(x)$ are the network's regression and softmax outputs parametrized by $\omega_t$~\cite{kendall2017uncertainties}.
\begin{align}
E_{p(y|x,Y,X)}(y^r) & \approx \int f_\omega(x) q_\theta(\omega) d\omega \approx \frac{1}{T} \sum_{t=1}^T f_{\omega_t}(x) =: \hat{f}_\theta(x) \label{eq:exp} \\
p(y^c|x,Y,X) & \approx \int p_\omega(x) q_\theta(\omega) d\omega \approx \frac{1}{T} \sum_{t=1}^T p_{\omega_t}(x) =: \hat{p}_\theta(x) \label{eq:post}
\end{align}
We estimate uncertainties through the predictive variance $Var_{p(y|x,Y,X)}$. Kendall et al.~\cite{kendall2017uncertainties} argue that predictive uncertainty can be divided into aleatoric (data) uncertainty, which originates from missing information in the data, and epistemic (model) uncertainty, which is caused by the model's lack of knowledge about the data. Intuitively, uncertainty regarding future events mainly corresponds to aleatoric uncertainty but likely induces model variation (epistemic) as well.

For regression, we follow Kendall et al.'s approach to estimate epistemic uncertainty (Eq. \ref{eq:reg}). Computing the predictive variance over parameter variation captures noise in the model. We omit aleatoric uncertainty for regression, since it was not effective for our task. For classification variables, we follow Kwon et al.~\cite{kwon2020uncertainty} (Eq. \ref{eq:cls}). The epistemic term captures noise in the model by estimating variance over parameter samples, while the variance over the multinomial softmax distribution in the aleatoric term captures inherent sample-independent noise. In the classification case, uncertainties are averaged over all classes.
\begin{align}
Var_{p(y|x,Y,X)}(y^r) & \approx
\underbrace{\frac{1}{T} \sum^T_{t=1} (f_{\omega_t}(x) - \hat{f}_\theta(x))^2}_{\sigma^2_{epistemic}}
+
\underbrace{\frac{1}{T} \sum^T_{t=1} \sigma^2_{\omega_t}(x)}_{(\sigma^2_{aleatoric})} \label{eq:reg} \\
Var_{p(y|x,Y,X)}(y^c) & \approx
\underbrace{\frac{1}{T} \sum^T_{t=1} (p_{\omega_t}(x) - \hat{p}_\theta(x))^2}_{\sigma^2_{epistemic}}
+
\underbrace{\frac{1}{T} \sum^T_{t=1} p_{\omega_t}(x)(1-p_{\omega_t}(x))}_{\sigma^2_{aleatoric}} \label{eq:cls}
\end{align}
% \\
% H_{p(y|x,Y,X)}(y^c) & \approx -\sum_{k=1}^K \hat{p}_\theta^k(x) \log \hat{p}_\theta^k(x)
\subsection{Model, Data \& Training}
The model (suppl. material) consists of a Bayesian AlexNet-style Convolutional Network~\cite{krizhevsky2012imagenet} and a Bayesian Long Short-Term Memory network (LSTM)~\cite{hochreiter1997long}. We sample dropout masks with a dropout rate of $20\%$ once per video and per sample and reuse the same masks at each time step as proposed by Gal et al.~\cite{gal2016theoretically} for recurrent architectures. The AlexNet backbone has proven effective in a similar setting~\cite{bodenstedt2019active} and empirically gave the best results. State-of-the-art architectures such as ResNet~\cite{he2016deep} performed poorly as they appeared to learn from future frames through batch statistics in batch-normalization layers~\cite{ioffe2015batch}. Further, the AlexNet can be trained from scratch, which is beneficial for introducing dropout layers. The code is published on \url{https://www.gitlab.com/nct_tso_public/ins_ant}.

We train on the Cholec80 dataset~\cite{twinanda2016endonet} of 80 recorded cholecystectomies and anticipate $K=5$ sparsely used instruments which are associated with specific tasks in the surgical workflow, i.e. bipolar (appears in $4.8\%$ of frames), scissors ($1.8\%$), clipper ($3.2	\%$), irrigator ($5.3\%$), and specimen bag ($6.2\%$).  Grasper and hook are dropped as they are used almost constantly during procedures and hence, are not of interest for anticipation. We extract frames at $1$ fps, resize to width and height of $384\times 216$, process batches of $128$ sequential frames and accumulate gradients over three batches. We use 60 videos for training and 20 for testing. We train for 100 epochs using the Adam optimizer (learning rate $10^{-4}$). The loss~(Eq. \ref{eq:loss}) is composed of smooth L1~\cite{twinanda2018rsdnet} for the primary regression task, cross entropy (CE) for the regularizing classification task and L2-regularization, where $\theta$ are parameter estimates of the approximate distribution $q_ \theta(\omega)$. We set $\lambda = 10^{-2}$ and $\gamma = 10^{-5}$.

\begin{equation} \label{eq:loss}
L_h(x,\tau) = \sum_\tau (\text{Smooth}L1(\hat{f}^\tau_\theta(x),r_h(x,\tau)) + \lambda \cdot CE(\hat{p}^\tau_\theta(x),c_h(x,\tau))) + \gamma \cdot \|\theta\|^2_2
\end{equation}

\begin{table}
\centering
\caption{Comparison to baselines. Results for \textit{OracleHist} are parenthesized since it is an offline approach. We report the mean over instrument types in minutes per metric.}\label{tab:overview}
\begin{tabular}{|c|cc|cc|cc|cc|}
\hline
& \multicolumn{2}{|c|}{$h$ = 2 min.} & \multicolumn{2}{|c|}{$h$ = 3 min.} & \multicolumn{2}{|c|}{$h$ = 5 min.}& \multicolumn{2}{|c|}{$h$ = 7 min.}\\
\hline
& wMAE & \emph{pMAE}\footref{foot:pmae} & wMAE & \emph{pMAE}\footref{foot:pmae} & wMAE & \emph{pMAE}\footref{foot:pmae} & wMAE & \emph{pMAE}\footref{foot:pmae}\\
\hline
MeanHist & 0.56 & 0.93 & 0.85 & 1.34 & 1.41 & 2.14 & 1.97 & 2.79 \\
OracleHist (offline) & (0.49) & (0.83) & (0.71) & (1.18) & (1.11) & (1.73) & \textbf{(1.48)} & (2.23) \\
\hline
Ours non-Bayes. & 0.44 & 0.67 & \textbf{0.64} & 0.93 & 1.13 & 1.64 & 1.58 & 2.21 \\
Ours+Phase & \textbf{0.42} & 0.64 & 0.67 & 0.94 & \textbf{1.07} & 1.49 & 1.61 & 2.16 \\
Ours & 0.44 & 0.65 & \textbf{0.64} & 0.92 & 1.09 & 1.53 & 1.58 & 2.16 \\
\hline
\end{tabular}
\end{table}

\begin{table}
\centering
\caption{Instrument-wise error in minutes for a horizon of 3 minutes.}\label{tab:instr}
\resizebox{\textwidth}{!}{
\begin{tabular}{|c|cc|cc|cc|cc|cc|}
\hline
& \multicolumn{2}{|c|}{Bipolar} & \multicolumn{2}{|c|}{Scissors} & \multicolumn{2}{|c|}{Clipper} & \multicolumn{2}{|c|}{Irrigator} & \multicolumn{2}{|c|}{Specimen Bag} \\
\hline
& wMAE & \emph{pMAE}\footref{foot:pmae} & wMAE & \emph{pMAE}\footref{foot:pmae} & wMAE & \emph{pMAE}\footref{foot:pmae} & wMAE & \emph{pMAE}\footref{foot:pmae} & wMAE & \emph{pMAE}\footref{foot:pmae} \\
\hline
MeanH. & 0.85 & 1.36 & 0.81 & 1.42 & 0.80 & 1.29 & 0.89 & 1.40 & 0.89 & 1.24 \\
OracleH. & (0.79) & (1.17) & (0.76) & (1.32) & (0.80) & (1.39) & (0.73) & (1.11) & \textbf{(0.50)} & (0.95) \\
\hline
O. non-B. & \textbf{0.75} & 0.92 & \textbf{0.49} & 0.78 & 0.74 & 1.01 & 0.71 & 1.05 & 0.54 & 0.92 \\
O.+Ph. & 0.79 & 0.96 & 0.53 & 0.82 & 0.77 & 1.01 & 0.72 & 1.05 & 0.56 & 0.88 \\
Ours & 0.76 & 0.96 & 0.51 & 0.76 & \textbf{0.71} & 0.90 & \textbf{0.70} & 1.03 & 0.55 & 0.93 \\
\hline
\end{tabular}
}
\end{table}

\section{Evaluation}
\subsection{Anticipation Results}
\subsubsection{Evaluation Metrics:} We evaluate frame-wise based on a weighted mean absolute error (\textit{wMAE}). We average the MAE of 'anticipating' frames ($0 < r_h(x,\tau) < h$) and 'background' frames ($r_h(x,\tau)=h$) to compensate for the imbalance in the data. As instruments are not always predictable, a low recall does not necessarily indicate poor performance, making precision metrics popular for anticipation~\cite{gao2017red,vondrick2016anticipating}. We capture the idea of precision in the \textit{pMAE}\footnote[1]{\label{foot:pmae}\emph	{Remark:} For future research, we do not recommend using the \emph{pMAE} metric since it can have undesirable properties in some situations. Specifically, adding small errors outside of the horizon can improve the overall score since errors inside the horizon are typically higher. Nevertheless, anticipation performance is still well represented through the \emph{wMAE} metric, which captures errors both inside and outside of the horizon.} as the MAE of predictions with $0.1h < \hat{f}^\tau_\theta(x) < 0.9h$ when the model is anticipating $\tau$. Factors are chosen for robustness against variations during 'background' predictions.

\subsubsection{Baselines:}
\begin{figure}
\centering
%\frame{
\includegraphics[width=\textwidth,trim=1.5cm 3.5cm .3cm 5.5cm, clip]{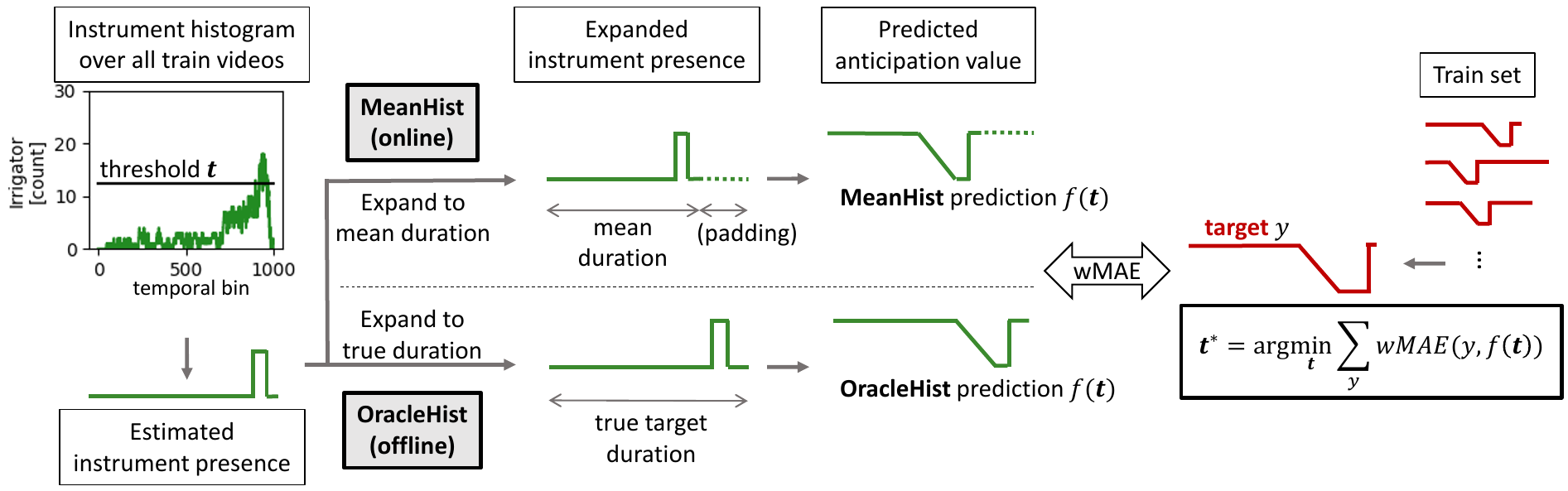}
%}
\caption{Overview of the baseline construction (example: irrigator). The temporal histogram of instrument occurences is thresholded to obtain an estimated instrument presence. The instrument presence signal is used to generate anticipation values according to Eq. \ref{eq:definition}. For each instrument, we find the optimal threshold $t^*$ w.r.t. \emph{wMAE} on the train set. For the online method \emph{MeanHist}, predictions are expanded to the mean duration of procedures in the train set. For the offline method \emph{OracleHist}, we assume the length of the target procedure to be known and expand predictions accordingly.}\label{fig:baseline}
\end{figure}

Since our task is not comparable to current anticipation methods, we compare to two histogram-based baselines.
For instrument $\tau$, the $i^{th}$ bin accumulates the occurrences of $\tau$ within the $i^{th}$ segments of all training video.
If the bin count exceeds a learned threshold, we assume that $\tau$ occurs regularly in the $i^{th}$ video segment and generate anticipation values according to Eq. \ref{eq:definition}.
Using 1000 bins, the thresholds are optimized to achieve best training performance w.r.t our main metric wMAE.
For \textit{MeanHist}, segments are expanded to the mean video duration. For \textit{OracleHist}, we expand the segments to the real video duration at train and test time. This is a strong baseline as instrument usage correlates strongly with the progress of surgery, which is not known beforehand. See Fig.~\ref{fig:baseline} for a visual overview of the baseline construction.

Additionally, we compare our model to a variant simultaneously trained on dense surgical phase segmentations~\cite{twinanda2016endonet} to investigate whether our model achieves competitive results using only sparse instrument labels. Surgical phases strongly correlate with instrument usage~\cite{twinanda2016endonet} and have shown to be beneficial for the related tasks of predicting the remaining surgery duration~\cite{twinanda2018rsdnet}. Finally, we compare to a non-Bayesian variant of our model without dropout to show that the Bayesian setting does not lead to a decline in performance.

\begin{figure}
\centering
\includegraphics[width=\textwidth]{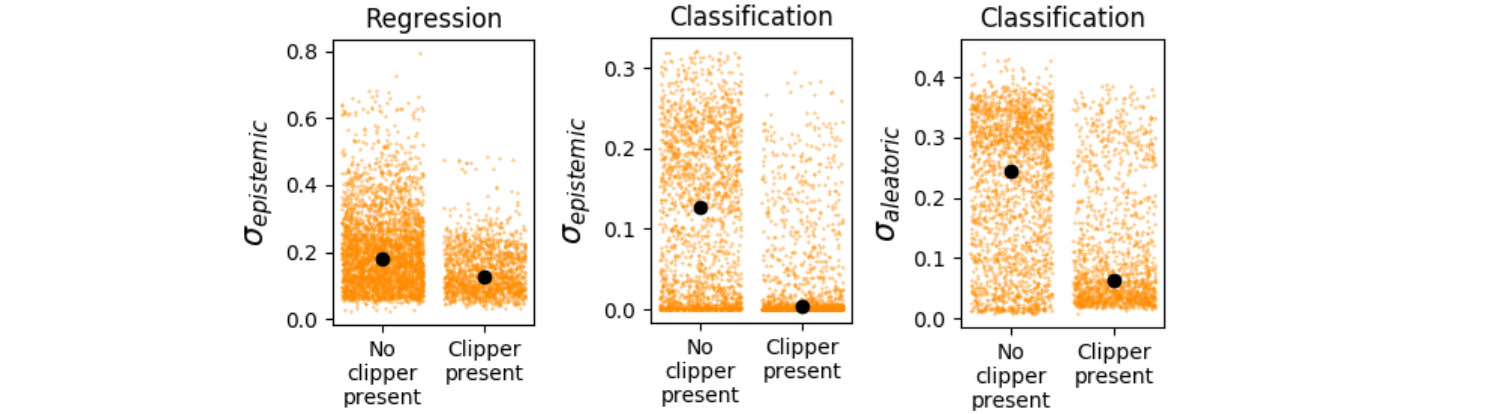}
\caption{Frame-wise (orange) and median (black) uncertainties for anticipating predictions of scissors depending on the clipper's presence. Since the clipper often proceeds scissors we expect lower uncertainty when the clipper is present.} \label{fig:scissors_clipper}
\end{figure}

\subsubsection{Discussion:} We train models on horizons of 2, 3, 5 and 7 minutes and repeat runs for \textit{Ours} and \textit{Ours non-Bayes.} four times and \textit{Ours+Phase} twice. In all settings, our methods outperform \textit{MeanHist} by a large margin (Table \ref{tab:overview}). Compared to the offline baseline \textit{OracleHist}, we achieve
%lower pMAE and comparable
slighty lower
wMAE errors, even though knowledge about the duration of a procedure provides strong information regarding the occurrence of instruments. Further, there is no visible difference in performance with and without surgical phase regularization. This suggests that our approach performs well by learning only from sparse instrument labels. The Bayesian setting also does not lead to a drop in performance while adding the advantage of uncertainty estimation. For $h=3$, we outperform \textit{MeanHist} for all instruments and \textit{OracleHist} for all except the specimen bag (Table \ref{tab:instr}). However, this instrument is easy to anticipate when the procedure duration is known since it is always used toward the end. For instrument-wise errors of other horizons, see the supplementary material.

\subsection{Uncertainty Quantification Results} \label{sec:uncert}
We analyze the model's ability to quantify uncertainties using a model for $h=3$. For all experiments, we consider predictions which indicate that the model is anticipating an instrument. We define \textit{anticipating predictions} as $\hat{f}^\tau_\theta(x)$ where $0.1\cdot h < \hat{f}^\tau_\theta(x) < 0.9\cdot h$, and $\hat{p}^\tau_\theta(x)$ with $\text{argmax}_j \hat{p}^\tau_\theta(x) = anticipating$.
\begin{figure}
\centering
\includegraphics[width=\textwidth]{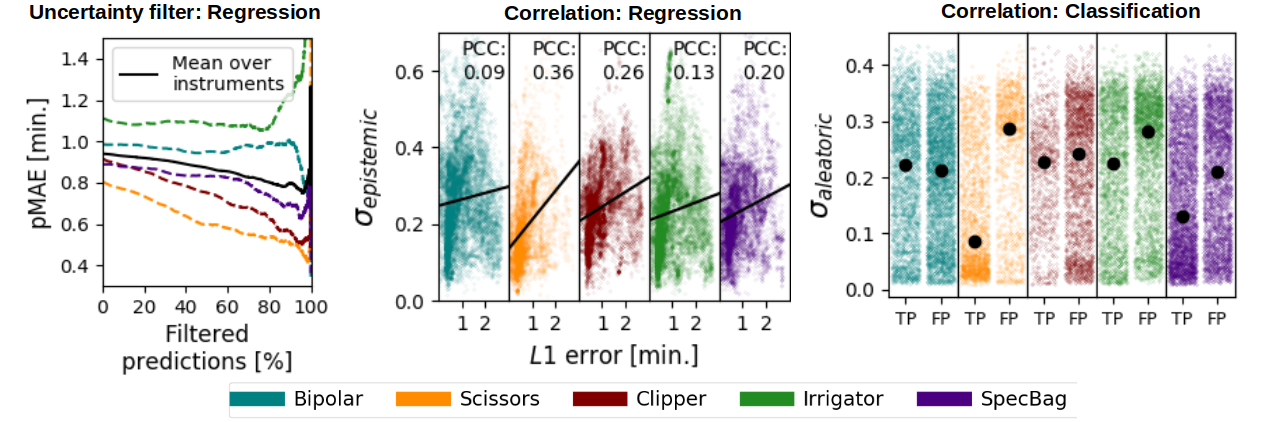}
\caption{\textit{Left:} pMAE as a plot of percentiles of filtered predictions w.r.t. $\sigma_{epistemic}$. \textit{Center:} Frame-wise error-uncertainty plot per instrument with Pearson Correlation Coefficient (PCC) and linear fit (black). \textit{Right:} Frame-wise (colored) and median (black) uncertainty of true positives (TP) vs. false positives (FP) for class \textit{anticipating}.}\label{fig:corr}
\end{figure}

Uncertainty quantification enables identification of events which trigger the usage of instruments. We evaluate the model using the known event-trigger relationship of scissors and clipper in cholecystectomies, where the cystic duct is first clipped and subsequently cut. Hence, we expect lower uncertainty for anticipating scissors when a clipper is visible. Fig. \ref{fig:scissors_clipper} supports this hypothesis for epistemic uncertainty during regression. However, the difference is marginal and most likely not sufficient for identifying trigger events. Even though clipper occurrence makes usage of the scissors foreseeable, predicting the exact time of occurrence is challenging and contains noise. Uncertainties for classification are more discriminative. The classification objective eliminates the need for exact timing and enables high-certainty class predictions. Both epistemic and aleatoric estimates are promising. This is consistent with our hypothesis that uncertainty regarding future events is contained in aleatoric uncertainty but induces epistemic uncertainty as well.

We assess the quality of uncertainty estimates through correlations with erroneous predictions as high uncertainty should result in higher errors. For regression (Fig. \ref{fig:corr}, center), we observe the highest Pearson Correlation Coefficients (PCC) for scissors, clipper and specimen bag. These instruments are presumably the most predictable since they empirically yield the best results (Table \ref{tab:instr}) and are correlated with specific surgical phases ('Clipping \& Cutting' and 'Gallbladder Packaging')~\cite{twinanda2016endonet}. Irrigator and Bipolar yield less reliable predictions as they are used more dynamically. For classification (Fig. \ref{fig:corr}, right), the median aleatoric uncertainty of true positive predictions for the 'anticipating' class is almost consistently lower than for false positives. Scissors and specimen bag show the largest margin. We can reduce precision errors (pMAE) by filtering uncertain predictions, shown in Fig. \ref{fig:corr} (left) for epistemic regression uncertainty. As expected, the decrease is steaper for instruments with higher PCC. See the supplementary material for corresponding plots with other horizons.

\section{Conclusion}
We propose a novel task and model for uncertainty-aware anticipation of intra-operative instrument usage. Limitations of current anticipation methods are addressed by enabling anticipation of sparse events in long videos. Our approach outperforms several baselines and matches a model variant using richer annotations, which indicates that sparse annotations suffice for this task. Since uncertainty estimation is useful for both anticipation tasks and intra-operative applications, we employ a probabilistic model. We demonstrate the model's ability to quantify task-relevant uncertainties by investigating error-uncertainty correlations and show that we can reduce errors by filtering uncertain predictions. Using a known example, we illustrate the model's potential for identifying trigger events for instruments, which could be useful for robotic applications.
Future work could investigate more effective methods for uncertainty quantification where aleatoric uncertainty is especially interesting due to its link to future events.

\bibliographystyle{splncs04}
\bibliography{mybibliography}

\begin{thebibliography}{10}
\providecommand{\url}[1]{\texttt{#1}}
\providecommand{\urlprefix}{URL }
\providecommand{\doi}[1]{https://doi.org/#1}

\bibitem{abu2019uncertainty}
Abu~Farha, Y., Gall, J.: Uncertainty-aware anticipation of activities. In:
  Proceedings of the IEEE International Conference on Computer Vision Workshops
  (2019)

\bibitem{abu2018will}
Abu~Farha, Y., Richard, A., Gall, J.: When will you do what?-anticipating
  temporal occurrences of activities. In: Proceedings of the IEEE Conference on
  Computer Vision and Pattern Recognition. pp. 5343--5352 (2018)

\bibitem{bodenstedt2019active}
Bodenstedt, S., Rivoir, D., Jenke, A., Wagner, M., Breucha, M.,
  M{\"u}ller-Stich, B., Mees, S.T., Weitz, J., Speidel, S.: Active learning
  using deep bayesian networks for surgical workflow analysis. International
  journal of computer assisted radiology and surgery  \textbf{14}(6),
  1079--1087 (2019)

\bibitem{damen2018scaling}
Damen, D., Doughty, H., Maria~Farinella, G., Fidler, S., Furnari, A., Kazakos,
  E., Moltisanti, D., Munro, J., Perrett, T., Price, W., et~al.: Scaling
  egocentric vision: The epic-kitchens dataset. In: Proceedings of the European
  Conference on Computer Vision (ECCV). pp. 720--736 (2018)

\bibitem{du2016recurrent}
Du, N., Dai, H., Trivedi, R., Upadhyay, U., Gomez-Rodriguez, M., Song, L.:
  Recurrent marked temporal point processes: Embedding event history to vector.
  In: Proceedings of the 22nd ACM SIGKDD International Conference on Knowledge
  Discovery and Data Mining. pp. 1555--1564 (2016)

\bibitem{gal2016dropout}
Gal, Y., Ghahramani, Z.: Dropout as a bayesian approximation: Representing
  model uncertainty in deep learning. In: international conference on machine
  learning. pp. 1050--1059 (2016)

\bibitem{gal2016theoretically}
Gal, Y., Ghahramani, Z.: A theoretically grounded application of dropout in
  recurrent neural networks. In: Advances in neural information processing
  systems. pp. 1019--1027 (2016)

\bibitem{gal2017deep}
Gal, Y., Islam, R., Ghahramani, Z.: Deep bayesian active learning with image
  data. In: Proceedings of the 34th International Conference on Machine
  Learning-Volume 70. pp. 1183--1192. JMLR. org (2017)

\bibitem{gao2017red}
Gao, J., Yang, Z., Nevatia, R.: Red: Reinforced encoder-decoder networks for
  action anticipation  (2017)

\bibitem{graves2011practical}
Graves, A.: Practical variational inference for neural networks. In: Advances
  in neural information processing systems. pp. 2348--2356 (2011)

\bibitem{he2016deep}
He, K., Zhang, X., Ren, S., Sun, J.: Deep residual learning for image
  recognition. In: Proceedings of the IEEE conference on computer vision and
  pattern recognition. pp. 770--778 (2016)

\bibitem{hochreiter1997long}
Hochreiter, S., Schmidhuber, J.: Long short-term memory. Neural computation
  \textbf{9},  1735--80 (12 1997). \doi{10.1162/neco.1997.9.8.1735}

\bibitem{ioffe2015batch}
Ioffe, S., Szegedy, C.: Batch normalization: Accelerating deep network training
  by reducing internal covariate shift. In: International Conference on Machine
  Learning. pp. 448--456 (2015)

\bibitem{jain2016recurrent}
Jain, A., Singh, A., Koppula, H.S., Soh, S., Saxena, A.: Recurrent neural
  networks for driver activity anticipation via sensory-fusion architecture.
  In: 2016 IEEE International Conference on Robotics and Automation (ICRA). pp.
  3118--3125. IEEE (2016)

\bibitem{ke2019time}
Ke, Q., Fritz, M., Schiele, B.: Time-conditioned action anticipation in one
  shot. In: Proceedings of the IEEE Conference on Computer Vision and Pattern
  Recognition. pp. 9925--9934 (2019)

\bibitem{kendall2017uncertainties}
Kendall, A., Gal, Y.: What uncertainties do we need in bayesian deep learning
  for computer vision? In: Advances in neural information processing systems.
  pp. 5574--5584 (2017)

\bibitem{krizhevsky2012imagenet}
Krizhevsky, A., Sutskever, I., Hinton, G.E.: Imagenet classification with deep
  convolutional neural networks. In: Advances in neural information processing
  systems. pp. 1097--1105 (2012)

\bibitem{kwon2020uncertainty}
Kwon, Y., Won, J.H., Kim, B.J., Paik, M.C.: Uncertainty quantification using
  bayesian neural networks in classification: Application to biomedical image
  segmentation. Computational Statistics \& Data Analysis  \textbf{142},
  106816 (2020)

\bibitem{mahmud2017joint}
Mahmud, T., Hasan, M., Roy-Chowdhury, A.K.: Joint prediction of activity labels
  and starting times in untrimmed videos. In: Proceedings of the IEEE
  International Conference on Computer Vision. pp. 5773--5782 (2017)

\bibitem{maier2014can}
Maier-Hein, L., Mersmann, S., Kondermann, D., Bodenstedt, S., Sanchez, A.,
  Stock, C., Kenngott, H.G., Eisenmann, M., Speidel, S.: Can masses of
  non-experts train highly accurate image classifiers? In: International
  conference on medical image computing and computer-assisted intervention. pp.
  438--445. Springer (2014)

\bibitem{maier2017surgical}
Maier-Hein, L., Vedula, S.S., Speidel, S., Navab, N., Kikinis, R., Park, A.,
  Eisenmann, M., Feussner, H., Forestier, G., Giannarou, S., et~al.: Surgical
  data science for next-generation interventions. Nature Biomedical Engineering
   \textbf{1}(9),  691--696 (2017)

\bibitem{mehrasa2019variational}
Mehrasa, N., Jyothi, A.A., Durand, T., He, J., Sigal, L., Mori, G.: A
  variational auto-encoder model for stochastic point processes. In:
  Proceedings of the IEEE Conference on Computer Vision and Pattern
  Recognition. pp. 3165--3174 (2019)

\bibitem{shridhar2018uncertainty}
Shridhar, K., Laumann, F., Liwicki, M.: Uncertainty estimations by softplus
  normalization in bayesian convolutional neural networks with variational
  inference. arXiv preprint arXiv:1806.05978  (2018)

\bibitem{twinanda2016endonet}
Twinanda, A.P., Shehata, S., Mutter, D., Marescaux, J., De~Mathelin, M., Padoy,
  N.: Endonet: A deep architecture for recognition tasks on laparoscopic
  videos. IEEE transactions on medical imaging  \textbf{36}(1),  86--97 (2016)

\bibitem{twinanda2018rsdnet}
Twinanda, A.P., Yengera, G., Mutter, D., Marescaux, J., Padoy, N.: Rsdnet:
  Learning to predict remaining surgery duration from laparoscopic videos
  without manual annotations. IEEE transactions on medical imaging
  \textbf{38}(4),  1069--1078 (2018)

\bibitem{vondrick2016anticipating}
Vondrick, C., Pirsiavash, H., Torralba, A.: Anticipating visual representations
  from unlabeled video. In: Proceedings of the IEEE Conference on Computer
  Vision and Pattern Recognition. pp. 98--106 (2016)

\bibitem{wang2019aleatoric}
Wang, G., Li, W., Aertsen, M., Deprest, J., Ourselin, S., Vercauteren, T.:
  Aleatoric uncertainty estimation with test-time augmentation for medical
  image segmentation with convolutional neural networks. Neurocomputing
  \textbf{338},  34--45 (2019)

\bibitem{zhong2018time}
Zhong, Y., Xu, B., Zhou, G.T., Bornn, L., Mori, G.: Time perception machine:
  Temporal point processes for the when, where and what of activity prediction.
  arXiv preprint arXiv:1808.04063  (2018)

\end{thebibliography}
\title{Rethinking Anticipation Tasks: Uncertainty-aware Anticipation of Sparse Surgical Instrument Usage for Context-aware Assistance (Supplementary Material)}
\titlerunning{Anticipation of Surgical Instruments}
\author{D. Rivoir et al.}
\institute{}
\authorrunning{D. Rivoir et al.}
\maketitle
\begin{figure}
\centering
%\frame{
\includegraphics[width=.94\textwidth,trim=5.5cm 9cm 3.5cm 3cm, clip]{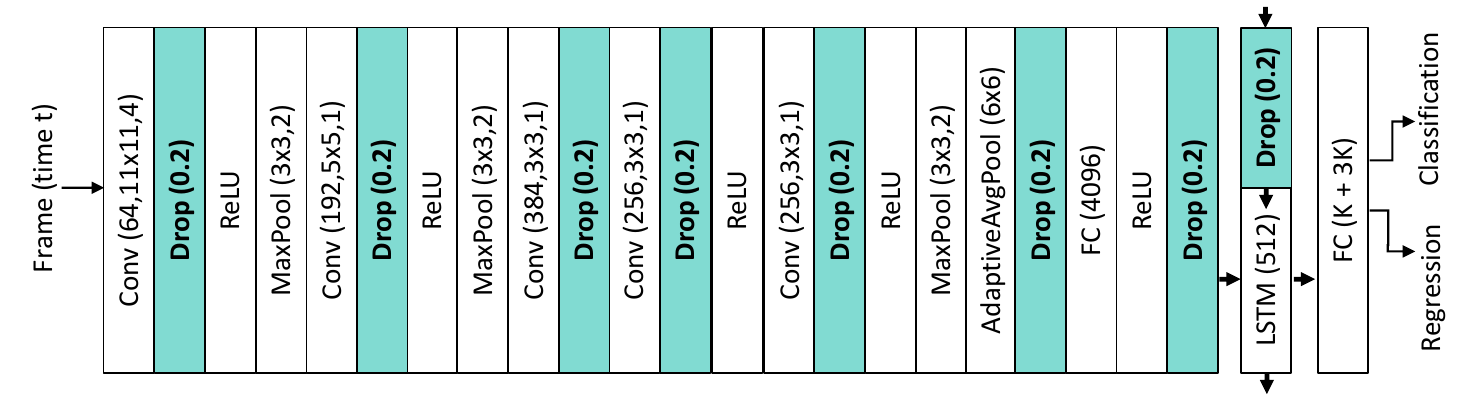}
%}
\caption{Network architecture for $K$ instruments: Dropout masks (\textbf{Drop}) are sampled once per video and reused at each time step. Notation: Conv (kernels, kernel size, stride), Drop (drop rate), MaxPool (kernel size, stride), AdaptiveAvgPool (output size), FC \textit{(fully-connected)} (size), LSTM (size). In the LSTM, dropout is applied to both the input and the hidden state of the previous time point.} \label{fig:network}
\end{figure}
\begin{table}[h]
\centering
\caption{Instrument-wise errors in minutes for horizons of 2, 5 and 7 min. The offline method \emph{OracleHist} is parenthesized.}
\begin{tabular}{|c|cc|cc|cc|cc|cc|}
\hline
& \multicolumn{2}{|c|}{Bipolar} & \multicolumn{2}{|c|}{Scissors} & \multicolumn{2}{|c|}{Clipper} & \multicolumn{2}{|c|}{Irrigator} & \multicolumn{2}{|c|}{Specimen Bag} \\
\hline
& wMAE & pMAE\footref{foot:pmae} & wMAE & pMAE & wMAE & pMAE & wMAE & pMAE & wMAE & pMAE \\
\hline
\multicolumn{11}{|c|}{$h$ = 2 min.} \\
\hline
MeanH. & 0.55 & 0.90 & 0.55 & 1.01 & 0.54 & 0.90 & 0.58 & 0.99 & 0.60 & 0.83 \\
OracleH. & (0.54) & (0.89) & (0.51) & (0.91) & (0.53) & (0.79) & (0.50) & (0.78) & (0.39) & (0.72) \\
\hline
O. non-b.& \textbf{0.51} & 0.71 & \textbf{0.33} & 0.57 & 0.50 &0.67 & 0.47 & 0.76 & 0.38 & 0.65 \\
O.+Ph. & \textbf{0.51} & 0.68 & \textbf{0.33} & 0.56 & \textbf{0.47} & 0.62 & \textbf{0.46} & 0.70 & \textbf{0.36} & 0.63 \\
Ours & \textbf{0.51} & 0.66 & 0.36 & 0.61 & 0.50 & 0.65 & 0.47 & 0.72 & 0.37 & \textbf{0.61} \\
\hline
\multicolumn{11}{|c|}{$h$ = 5 min.} \\
\hline
MeanH. & 1.44 & 2.26 & 1.33 & 2.18 & 1.34 & 2.12 & 1.50 & 2.15 & 1.46 & 1.98 \\
OracleH. & \textbf{(1.25)} & (1.67) & (1.31) & (2.20) & (1.23) & (1.93) & \textbf{(1.14)} & (1.63) & \textbf{(0.64)} & (1.28) \\
\hline
O. non-b.& 1.36 & 1.70 & 0.93 & 1.41 & 1.24 & 1.75 & 1.25 & 1.80 & 0.90 & 1.54 \\
O.+Ph. & 1.32 & 1.61 & \textbf{0.88} & 1.33 & \textbf{1.09} & 1.55 & 1.22 & 1.63 & 0.85 & 1.34 \\
Ours & 1.33 & 1.67 & 0.89 & 1.32 & 1.12 & 1.64 & 1.25 & 1.64 & 0.85 & 1.40 \\
\hline
\multicolumn{11}{|c|}{$h$ = 7 min.} \\
\hline
MeanH. & 2.08 & 2.90 & 1.82 & 2.76 & 1.86 & 2.93 & 2.14 & 2.78 & 1.95 & 2.59 \\
OracleH. & \textbf{(1.71)} & (2.18) & (1.78) & (2.90) & (1.64) & (2.39) & \textbf{(1.54)} & (2.16) & \textbf{(0.74)} & (1.50) \\
\hline
O. non-b.& 2.00 & 2.34 & 1.37 & 2.02 & 1.56 & 2.30 & 1.82 & 2.31 & 1.18 & 2.07 \\
O.+Ph. & 2.14 & 2.43 & 1.31 & 1.98 & 1.54 & 2.14 & 1.89 & 2.26 & 1.22 & 2.01 \\
Ours & 2.04 & 2.39 & \textbf{1.26} & 1.89 & \textbf{1.51} & 2.20 & 1.86 & 2.30 & 1.23 & 2.04 \\
\hline
\end{tabular}
\end{table}

\begin{figure}
\centering
\includegraphics[width=.88\textwidth]{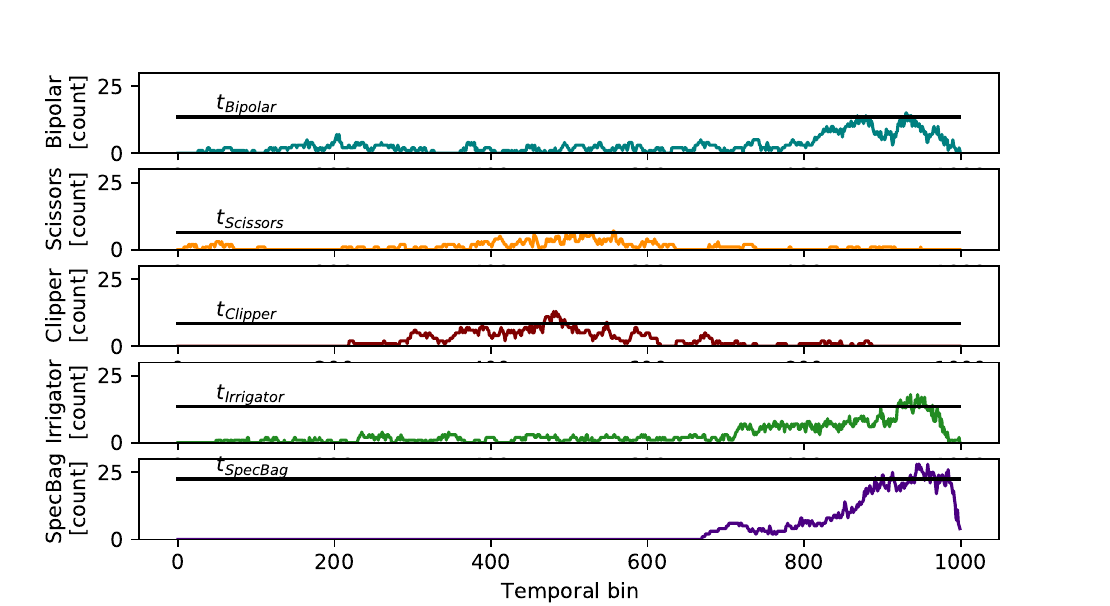}
\caption{Histogram of instrument occurrences over the training set used for learning the baselines \emph{MeanHist} and \emph{OracleHist}. Horizontal lines $t_{<instrument>}$ illustrate an example of learnt thresholds for \emph{OracleHist} and a horizon of 3 min.}
\end{figure}

\begin{figure}
\centering
%\frame{
\includegraphics[width=\textwidth,trim=3.5cm 1cm 4.9cm 2.5cm, clip]{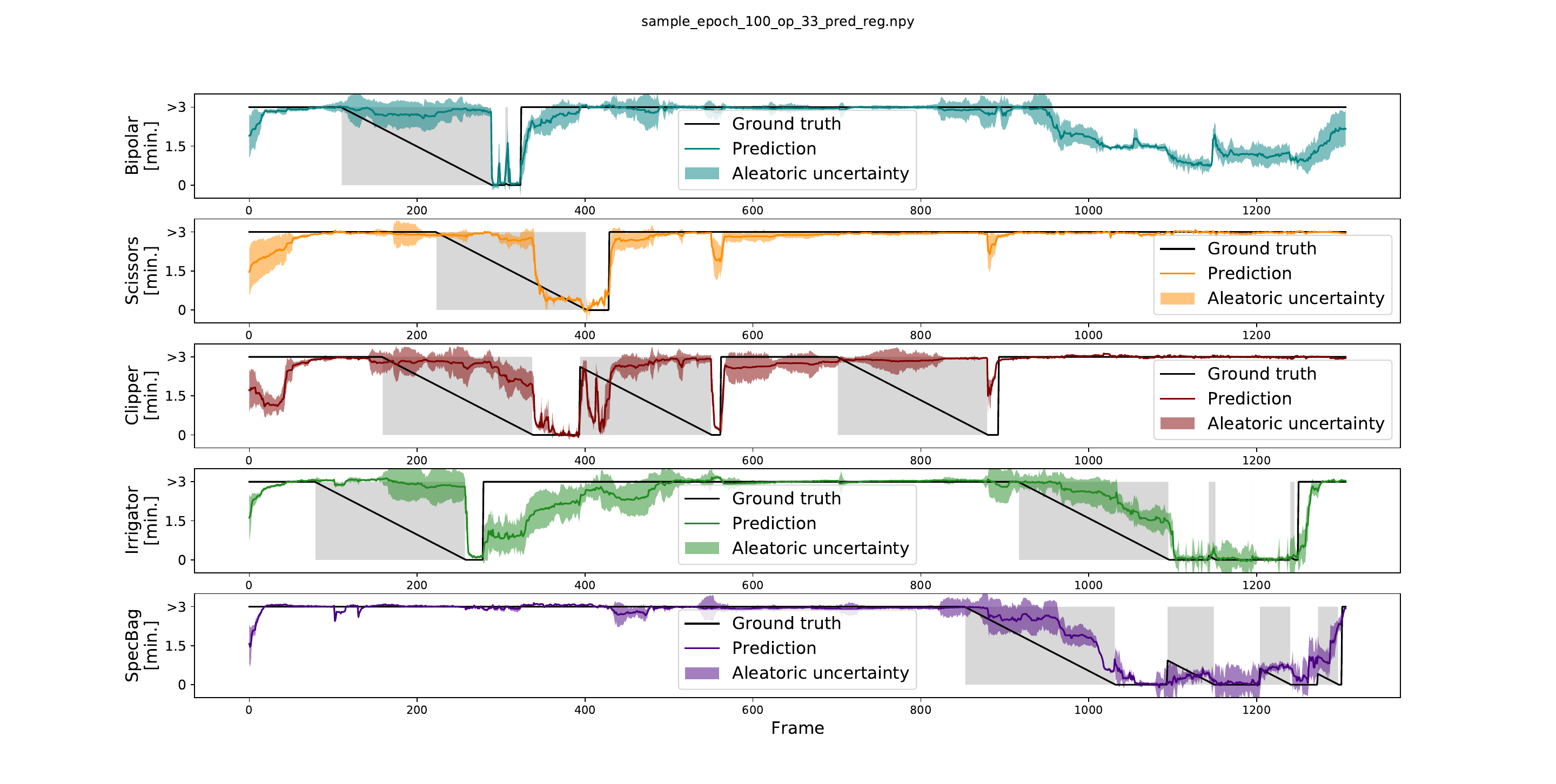}
%}
\caption{Anticipation results for example surgery with $h=3$. Grey areas highlight anticipating frames ($0<r_h(x,\tau)<h$). The best anticipation results are achieved for scissors and specimen bag. Our model anticipates the clipper when used directly before scissors but fails otherwise as it is used outside of the typical workflow of cholecystectomies.}
\end{figure}

\begin{figure}
\centering
%\frame{
\includegraphics[width=.7\textwidth,trim=3cm 24cm 0cm .5cm, clip]{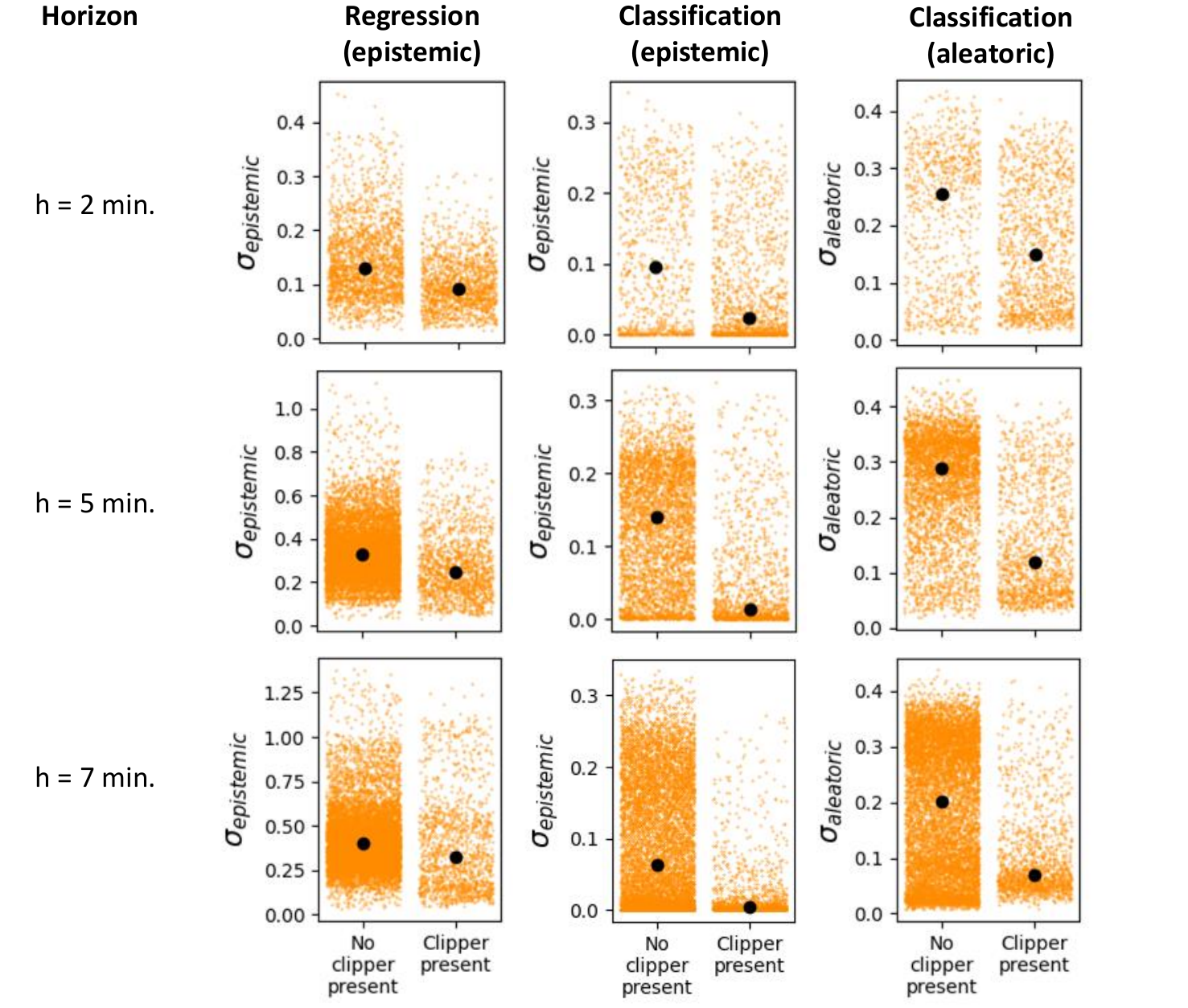}
%}
\caption{Uncertainty for anticpating scissors with horizons of 2, 5 and 7 minutes. The plot shows frame-wise (orange) and median (black) uncertainties for anticipating predictions of scissors depending on the clipper's presence. Since the clipper often proceeds scissors we expect lower uncertainty when the clipper is present.}
\end{figure}

\begin{figure}
\centering
%\frame{
\includegraphics[width=.97\textwidth,trim=0cm 22.3cm 0cm 1cm, clip]{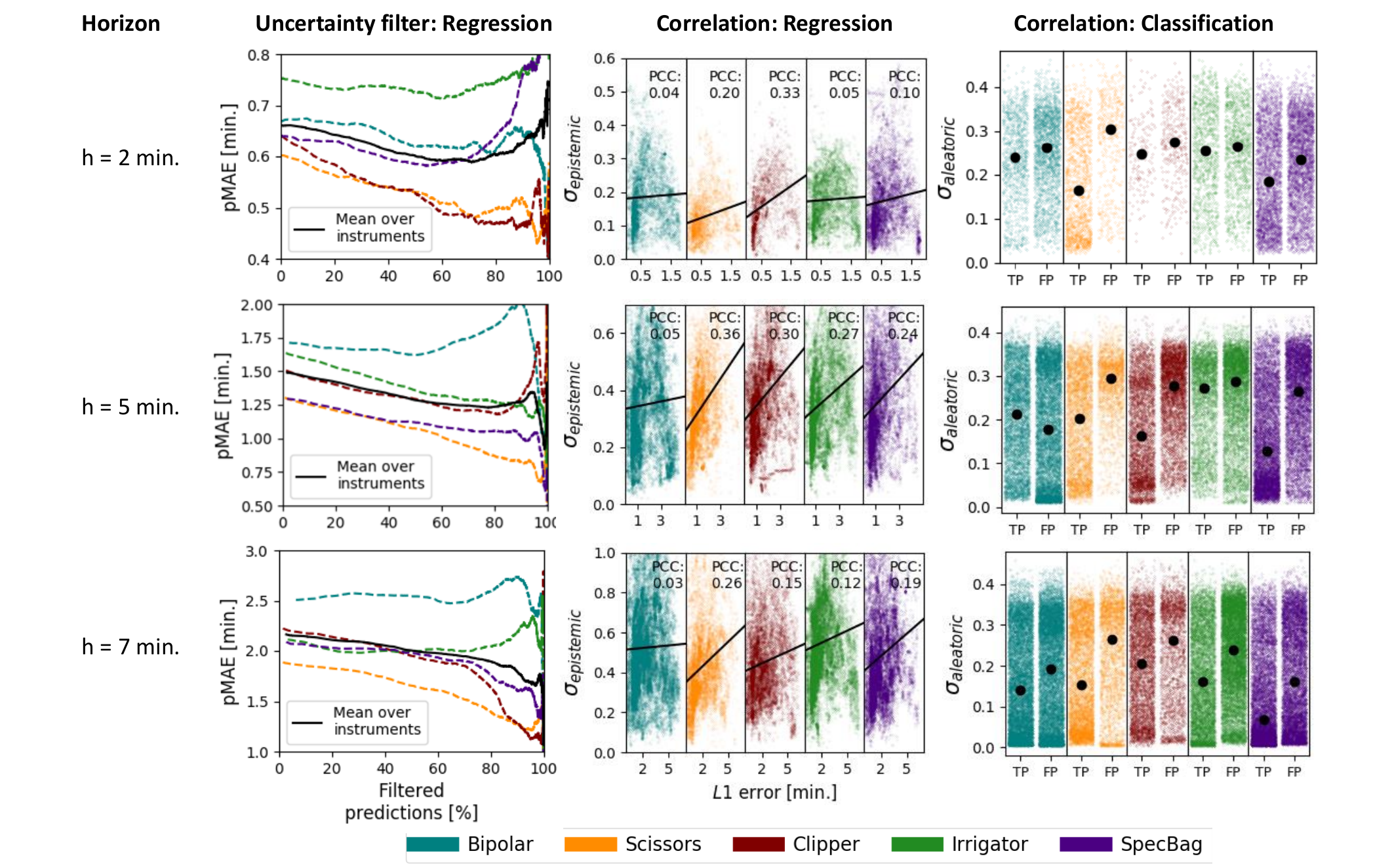}
%}
\caption{Error-uncertainty plots for horizons of 2, 5 and 7 minutes. \textit{Left:} pMAE as a plot of percentiles of filtered predictions w.r.t. $\sigma_{epistemic}$. \textit{Center:} Frame-wise error-uncertainty plot per instrument with Pearson Correlation Coefficient (PCC) and linear fit (black). \textit{Right:} Frame-wise (colored) and median (black) uncertainty of true positives (TP) vs. false positives (FP) for class \textit{anticipating}.}
\end{figure}

\end{document}